\documentclass[a4paper,11pt]{article}

\usepackage{breakurl}
\usepackage[breaklinks=true]{hyperref}
\usepackage{breakcites}

\usepackage{amssymb,amsmath,array,bm}
\usepackage{booktabs}

\usepackage{tikz}
\usetikzlibrary{arrows}

\usepackage{pgfplots}
\pgfplotsset{compat=1.15}
\usetikzlibrary{arrows}

\usepackage[a4paper, left=3cm, right=2.5cm, top=3cm, bottom=3cm]{geometry}

\usepackage{authblk}
\usepackage{natbib}
\usepackage{doi}

\usepackage[utf8]{inputenc}
\usepackage[T1]{fontenc}
\usepackage[english]{babel}
\usepackage{csquotes}

\title{Does Interpretability of Knowledge Tracing Models Support Teacher Decision Making?}

\author{Adia Khalid}
\author{Alina Deriyeva}
\author{Benjamin Paaßen}
\affil{Faculty of Technology, Bielefeld University}

\date{preprint as provided by the authors}

\begin{document}

\maketitle

\pagestyle{myheadings}
\markright{preprint as provided by the authors}

\begin{abstract}
Knowledge tracing (KT) models are a crucial basis for pedagogical decision-making, namely which task to select next for a learner and when to stop teaching a particular skill. Given the high stakes of pedagogical decisions, KT models are typically required to be interpretable, in the sense that they should implement an explicit model of human learning and provide explicit estimates of learners' abilities. However, to our knowledge, no study to date has investigated whether the interpretability of KT models actually helps human teachers to make teaching decisions. We address this gap. First, we perform a simulation study to show that, indeed, decisions based on interpretable KT models achieve mastery faster compared to decisions based on a non-interpretable model. Second, we repeat the study but ask $N=12$ human teachers to make the teaching decisions based on the information provided by KT models. As expected, teachers rate interpretable KT models higher in terms of usability and trustworthiness. However, the number of tasks needed until mastery hardly differs between KT models. This suggests that the relationship between model interpretability and teacher decisions is not straightforward: teachers do not solely rely on KT models to make decisions and further research is needed to investigate how learners and teachers actually understand and use KT models.
\end{abstract}

\section{Introduction}

Knowledge tracing (KT) models are crucial parts of adaptive learning environments \cite{abdelrahman2023knowledge}. They predict how likely a student is to succeed in any of the possible next learning tasks and thereby help to decide which next task may be most suitable for a given student, i.e.\ the outer loop of an intelligent tutoring system \cite{vanlehn2006behavior}. Given the importance of pedagogical decisions, it has been argued that it is crucial for KT models to be \emph{interpretable} to students and teachers \cite{bull2010,chen2023improving,lu2020towards}. More specifically, deep KT models have been criticized for using excessive numbers of parameters and failing to provide explicit estimates of student ability \cite{khajah2016deep}. However, to our knowledge, no study to date has investigated how much model interpretability actually helps human teachers to make pedagogical decisions. In this paper, we aim to address this gap.

We perform two studies to investigate our research question: First, a simulation study in which we automatically select tasks to maximize the expected student learning gain based on different KT models; and we evaluate how fast (simulated) students learn based on the selected tasks. Second, we repeat the study but let $N=12$ teachers select the tasks and decide when to stop teaching, based on the information provided by the different KT models. The students are simulated via an Elo model and the KT models are Bayesian Knowledge Tracing (BKT) \cite{yudelson2013individualized}, Performance Factors Analysis (PFA) \cite{pavlik2009performance}, and Deep Knowledge Tracing (DKT) \cite{piech2015dkt}. BKT and PFA are interpretable in the sense that they have a well-defined underlying model of student learning with few parameters and offer explicit ability estimates, whereas DKT learns an implicit model using a recurrent neural network that does not have an immediate interpretation.  Our core research hypothesis is that simulated students learn faster when decisions are made based on BKT and PFA instead of DKT.

The main advantage of running our study in a simulation environment is that we have control over many variables that could influence teaching: We control exactly how our simulated students react to teacher decisions and learn, can mathematically compute which decisions would be optimal and can compare this to teachers' actual decisions. The only variable that changes between conditions is the KT model, nothing else. This does not replace research with human students but can isolate the specific effect of KT model interpretability and thus provide some insight if intepretability by itself helps teachers to make decisions.

Our contributions are: a) a simulation environment to run studies with simulated students, b) a simulation study including mathematical predictions of learning gains for BKT, PFA, and DKT, and c) the first user study with teachers to compare teacher decision-making based on the information provided by BKT, PFA, and DKT.

The experimental code can be found at \url{https://gitlab.ub.uni-bielefeld.de/publications-ag-kml/kt_space_tutor}.

\section{Background and Related Work}\label{sec:related_work}

In this work, we use a broad definition of knowledge tracing (KT) models, akin to \cite{abdelrahman2023knowledge}. For us, KT is defined by taking a sequence of previous attempts of a learner as input and outputting a success prediction of said learner for every possible next task. More formally, let $j_1, \ldots, j_T \in \{1, \ldots, m\}$ be a sequence of tasks a certain learner has attempted and let $x_1, \ldots, x_T \in \{0, 1\}$ be the corresponding sequence of success information -- $1$ represents success, $0$ represents failure. Then, we define a KT model as any function that maps a sequence $(j_1, x_1), \ldots, (j_T, x_T)$ to an $m$-dimensional vector of probabilities $\vec p_{t+1}$ where $p_{t+1, j}$ is the model's predicted probability that the student would successfully complete task $j$ in the next time step; that is, $x_{t+1}$ is predicted as $1$ with probability $p_{t+1, j}$ if $j_t+1 = j$. We further call a KT model \emph{interpretable}, if it provides an explicit ability estimate $\theta_t$ for each skill and each student in each time step. We now discuss all KT models considered in this paper in turn.

\paragraph{The Elo model} is commonly used in sports (particularly chess) to model skill development over time \cite{pelanek2016applications,ruettgers2024matchmaking}. Given a slope parameter $a > 0$, a difficulty parameter for each task $b_j$, and a learning rate parameter $\kappa > 0$, Elo is defined by the equations:
\begin{align}
p_{t, j} &= \frac{1}{1 + \exp(-a \cdot (\theta_t-b_j))} \text{, with }
\theta_{t+1} = \theta_t + \kappa \cdot (x_t - p_{t, j_t}),\label{eq:elo_update}
\end{align}
Because the parameters of the model are difficult to estimate from data, we do not use it for KT in this paper, but rather to simulate students. This way, we ensure that the underlying simulation is substantially different from any KT model and avoid giving any KT model an unfair advantage.

\paragraph{Bayesian Knowledge Tracing (BKT)} is a two-state Hidden Markov Model \cite{yudelson2013individualized}. In particular, let $X_t$ be a random variable modeling the success/failure of a student at the task at time $t$ ($X_t = 0$ for failure, $X_t = 1$ for success) and let $Z_t$ be a random variable modeling whether the student mastered the underlying skill at time $t$ ($Z_t = 0$ for not mastered and $Z_t = 1$ for mastered). Then, a BKT model is fully defined by the equations $p_{Z_1}(1) = p_\text{start}$, $p_{Z_t|Z_{t-1}}(1|1) = 1$, $p_{Z_T|Z_{t-1}}(1|0) = p_\text{trans}$, $p_{X_t|Z_t}(1|0) = p_\text{guess}$, and $p_{X_t|Z_t}(0|1) = p_\text{slip}$, where $p_\text{start}$, $p_\text{trans}$, $p_\text{guess}$, and $p_\text{slip}$ are the parameters of the model. All other probabilities follow via Bayesian reasoning. In our scenario, we are particularly interested in the probability $p_{Z_{t+1}|X_1, \ldots, X_t}(1|x_1, \ldots, x_t)$, meaning the probability that the student has mastered the skill in the next time step, given all their previous successes and failures. This is our BKT ability estimate $\theta_{t, k}$. Via Bayesian reasoning, we obtain the following recursive update formula for $\theta_{t, k}$:
\begin{align}
\theta_{t+1, k} = q_{t, k} + p_\text{trans} \cdot q_{t, k}\text{, with } 
q_{t, k} &= 
\begin{cases}
\frac{(1 - p_\text{slip}) \cdot \theta_{t, k}}{(1 - p_\text{slip}) \cdot \theta_{t, k} + p_\text{guess} \cdot (1 - \theta_{t, k})} & \text{if } x_t = 1 \\
\frac{p_\text{slip} \cdot \theta_{t, k}}{p_\text{slip} \cdot \theta_{t, k} + (1-p_\text{guess}) \cdot (1 - \theta_{t, k})} & \text{if } x_t = 0
\end{cases}. \label{eq:bkt_update}
\end{align}

\paragraph{Performance Factors Analysis (PFA)} is a logistic model \cite{pavlik2009performance}. In particular, let $\mathcal{K}(j)$ be the set of skills required by task $j$. Then, the predicted probability of success according to PFA is:
\begin{align}
p_{t+1, j} &= \frac{1}{1 + \exp(-z_{t+1, j})} \text{, with } 
z_{t+1, j} = \sum_{k \in \mathcal{K}(j)} \beta_k + \gamma_k \cdot s_{k, t} + \rho_k \cdot f_{k, t}, \label{eq:pfa}
\end{align}
where $s_{k, t} = \sum_{t : k \in \mathcal{K}(j_t)} x_t$ is the number of past successes of the student in skill $k$, $f_{k, t} = \sum_{t : k \in \mathcal{K}(j_t)} 1 - x_t$ is the number of past failures of the student in skill $k$, and $\beta_k$, $\gamma_k$, and $\rho_k$ are the parameters of the model, shared across students and tasks. $\beta_k$ can also be interpreted as the estimated initial ability of each student in skill $k$, $\gamma_k$ as the learning rate for successes in skill $k$, and $\rho_k$ as the learning rate for failures in skill $k$. Accordingly, we can interpret $\theta_{k, t} = \beta_k + \gamma_k \cdot s_{k, t} + \rho_k \cdot f_{k, t}$ as the estimated ability of the student in skill $k$ at time $t$. 

\paragraph{Deep Knowledge Tracing (DKT)} predicts the probability of future success using a recurrent neural network \cite{piech2015dkt}. In particular, every tuple $(j_t, x_t)$ is encoded as a $2m$-dimensional vector $\hat x_t$, which is zero everywhere except for $\hat x_{t, j_t} = 1$ if $x_t = 1$ and $\hat x_{t, j_t+m} = 1$ if $x_t = 0$. Then, the recurrent neural net maps these encodings to a latent state vector $\vec h_{t+1}$ and a feedforward neural net finally maps $h_t$ to the predicted success probabilities $\vec p_{t+1}$.  The state vector is updated via a recurrent neural network (a GRU in our case \cite{chung2014gru}), and the feedforward layer typically uses a sigmoid nonlinarity, same as~\eqref{eq:pfa}.

Importantly, the latent state $\vec h_t$ has no intuitive interpretation and there is no immediate way to extract an ability estimate from $\vec h_t$. 
We do acknowledge that many attempts exist to make DKT more interpretable, such as \cite{chen2023improving,lu2020towards}, but our research question is precisely whether interpretability is necessary in the first place. Therefore, we regard the default version of DKT as the non-interpretable control condition.

\subsection{Prior studies of KT model interpretability}

To our knowledge, our study is the first human user study to compare the interpretability of KT models. Still, several prior reviews have contrasted the interpretability of different KT models on a computational and conceptual level. \cite{khajah2016deep} compares DKT and BKT in depth, acknowledging that default DKT is more accurate in predicting future task success compared to BKT, but showing that BKT can be extended to match DKT's performance, while retaining its interpretability advantage. \cite{ding2019deep,wang2023wrong} go into more detail on DKT, highlighting counterintuitive behavior where DKT success predictions can wildly fluctuate, meaning that $p_{t+1, j}$ is very different from $p_{t, j}$, e.g.\ due to overfitting to the training data. \cite{gervet2020deep} compare logistic models (like PFA), BKT, and DKT in terms of accuracy for small and big data sets and find that logistic models are most accurate on small data sets, whereas DKT performs best on big data sets.

\cite{scruggs2023well} evaluate how well the ability estimates of different KT models predict the post-test scores in a game-based learning environments and found that a very simple baseline tends to outperform the ability estimates provided by most KT models (even state-of-the-art ones). This suggests that the ability estimates provided by KT models may be predictive of future task performance, but not necessarily of post-test performance. \cite{mao2018deep} find that BKT outperforms DKT when predicting post-test scores in another setting.

Overall, past research suggests that (default) DKT lacks interpretability and has counter-intuitive properties (like fluctuation), whereas PFA and BKT may be less accurate but provide more interpretability and each prediction is fully justified by explicit model equations. Our research question is whether this difference in interpretability also influences teachers' decision making and subjective usability and trust.

\section{Simulation Study}

To study the effectiveness of different KT models (PFA, BKT, and DKT) in making pedagogical decisions, we first perform a simulation study.

\subsection{Method}
\label{sec:simulation_method}

\begin{figure}
\centering
\includegraphics[width=0.65\linewidth]{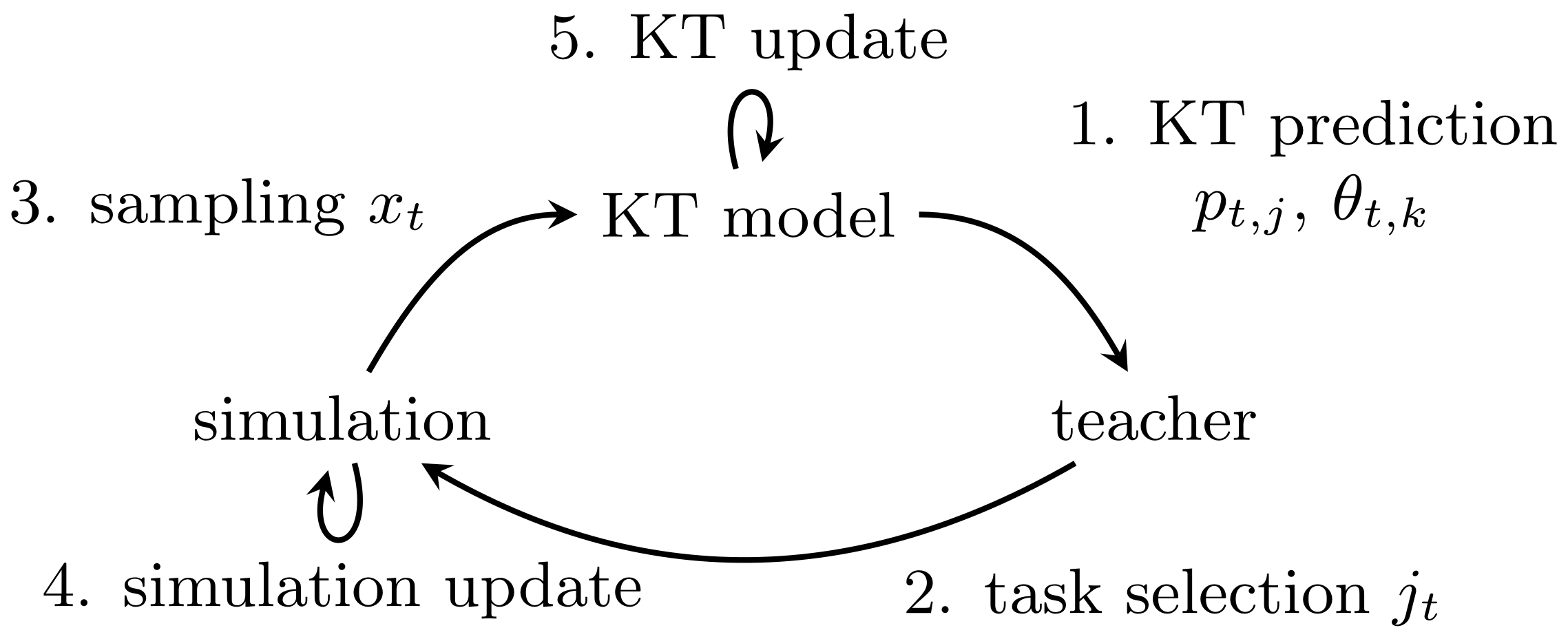}
1
\caption{The simulation loop involving knowledge tracing, task selection, sampling of success or failure, simulation update, and KT model update.}
\label{fig:simulation}
\end{figure}

We set up an Elo-based simulation environment that contains a minimal but non-trivial learning scenario. We include four learning tasks and two skills. Task 1 trains only skill 1, task 2 trains only skill 2, and tasks 3 and 4 train both skills. Tasks 1, 2, and 4 all have the difficulty parameter $b_j = 1$, but task 3 is easier with $b_3 = 0$. Then, we simulate each student using the scheme illustrated in Fig.~\ref{fig:simulation}. In particular, the steps of our method are:

\paragraph{Initialization:} We initialize one Elo model per skill, with initial Elo abilities $\theta^*_{1, 1} = \theta^*_{1, 2} = 0$. We denote the Elo abilities with an asterisk to distinguish them from the ability estimates of the KT models.

\paragraph{1. KT prediction:} Using the current KT model, we predict the success probability for each task $p_{t, j}$ and, if possible, an estimate $\theta_{t, k}$ for the student's current ability in each skill. These predictions and estimates follow from the equations presented in Section~\ref{sec:related_work}. Note that, for BKT, we run a separate BKT model for each skill and, if a task involved multiple skills, we take the smallest predicted success probability across all skills. For PFA, we add a difficulty parameter into the model to improve the match with our scenario \cite{Deriyeva2024ECAI}.

\paragraph{2. Task selection:} Using the current KT model, we select the next task $j_t$ for the student based on the highest expected learning gain. The expected learning gain can be quantified by summing over all skills involved in a task and adding the learning gain in case of success (according to the KT model) times the probability of success (according to the KT model) to the learning case in case of failure (according to the KT model) times the probability of failure (according to the KT model), minus the current ability estimate. For BKT, this yields the expression
\begin{align}
%	&\mathbb{E}[\Delta \theta^\text{BKT}_t(j)] = \\
&\sum_{k \in \mathcal{K}(j)} p_{t, j} \cdot \frac{(1 - p_\text{slip}) \cdot \theta_{t, k}}{(1 - p_\text{slip}) \cdot \theta_{t, k} + p_\text{guess} \cdot (1 - \theta_{t, k})} \notag \\
&+ (1-p_{t, j}) \cdot \frac{p_\text{slip} \cdot \theta_{t, k}}{p_\text{slip} \cdot \theta_{t, k} + (1-p_\text{guess}) \cdot (1 - \theta_{t, k})} - \theta_{t, k}. \label{eq:gain_bkt} 
\end{align}
For PFA, we gain $\gamma_k$ if the student succeeds and $\rho_k$ if the student fails, yielding the much simpler expression:
\begin{equation}
\sum_{k \in \mathcal{K}(j)} p_{t, j} \cdot \gamma_k + (1-p_{t, j}) \cdot \rho_k. \label{eq:gain_pfa}
\end{equation}
For DKT, the expected learning gain is particularly difficult to compute because there is no explicit ability estimate in DKT. The only proxy we have is the predicted probability of success for each task. In particular, let $\Delta p_{t, j'}^1(j)$ be the difference $p_{t+1, j'} - p_{t, j'}$ according to DKT if we feed another success on task $j$ into the DKT model, and let $\Delta p_{t, j'}^0(j)$ be the difference $p_{t+1, j'} - p_{t, j'}$ according to DKT if we feed another failure on task $j$ into the DKT model. Then, our notion of expected learning gain for DKT becomes:
\begin{equation}
\sum_{j' = 1}^m p_{t, j} \cdot \Delta p_{t, j'}^1(j) + (1-p_{t, j}) \cdot \Delta p_{t, j'}^0(j).  \label{eq:gain_dkt}
\end{equation}

\paragraph{3. Sampling:} Once a task $j_t$ is selected, we predict the success probability of the Elo model for each involved skill and then take the product over all skills involved in task $j_t$ to derive the success probability for the task. According to this probability, we then sample the success or failure $x_t$ of the simulated student. 

\paragraph{4. Simulation update:} Once the success or failure $x_t$ is sampled, we update the Elo model for each involved skill. Notice, though, that the standard Elo model would \emph{reduce} the skill estimate $\theta^*_{t, k}$ if $x_t = 0$. This is because, usually, the Elo model produces an \emph{estimate} of ability, which needs to be reduced upon receiving evidence for a lack of ability. By contrast, we use Elo as a ground truth for simulation -- and we assume that learners do not forget/lose ability during a single teaching session. Accordingly, we change the update formula~\eqref{eq:elo_update} as follows:
\begin{equation}
\theta^*_{t+1, k} = \theta^*_{t, k} +
\begin{cases}
\kappa_1 \cdot (1 - p^*_{t, k, j}) & \text{ if } k \in \mathcal{K}(j_t) \text{ and } x_t = 1 \\
\kappa_0 \cdot (1 - p^*_{t, k, j}) & \text{ if } k \in \mathcal{K}(j_t) \text{ and } x_t = 0 \\
0 & \text{ if } k \notin \mathcal{K}(j_t)
\end{cases}
\end{equation}
where $\kappa_1 = 1$ and $\kappa_0 = 0.5$ are different learning rates for successes and failures, respectively. These choices are made to ensure that the simulated students learn more (namely twice as much) from successes compared to failures.
In this setup, the \enquote{true} expected learning gain according to Elo would be:
\begin{equation}
\sum_{k \in \mathcal{K}(j)} p^*_{t, j} \cdot \kappa_1 \cdot (1 - p^*_{t, j, k}) 
+ (1-p^*_{t, j}) \cdot \kappa_0 \cdot (1 - p^*_{t, j, k}).
\end{equation}
Note that this strongly differs from the expected learning gain for all KT models (refer to \eqref{eq:gain_bkt}, \eqref{eq:gain_pfa}, and \eqref{eq:gain_dkt}). Therefore, no KT model has an a priori advantage.

\paragraph{5. KT update:} Based on the selected task $j_t$ and the sampled success/failure $x_t$, we update the KT model using the respective equations from Section~\ref{sec:related_work}. 

\paragraph{End of simulation:} Based on the updated KT model, we check whether the current KT model predicts mastery and would, therefore, end the teaching. We define mastery in our simulation as both skills exceeding ability $1.5$ (according to the Elo model). Therefore, in PFA, we predict mastery if all skill estimates exceed $1.5$. In BKT and DKT, we stop teaching if the predicted success probability across all tasks corresponds to an ability level of $1.5$ in a logistic model. If mastery is \emph{not} yet predicted, we increment the time step $t$ by one and return to step (1) of the simulation loop.

To generate training data for our KT models, we first simulate $500$ students with random task choices in step 3. Then, we train our KT models on this data and evaluate them by simulating $1000$ additional students per model.

\subsection{Results}
\label{sec:simulation_results}

\begin{figure}
\centering
\includegraphics[width=1\linewidth]{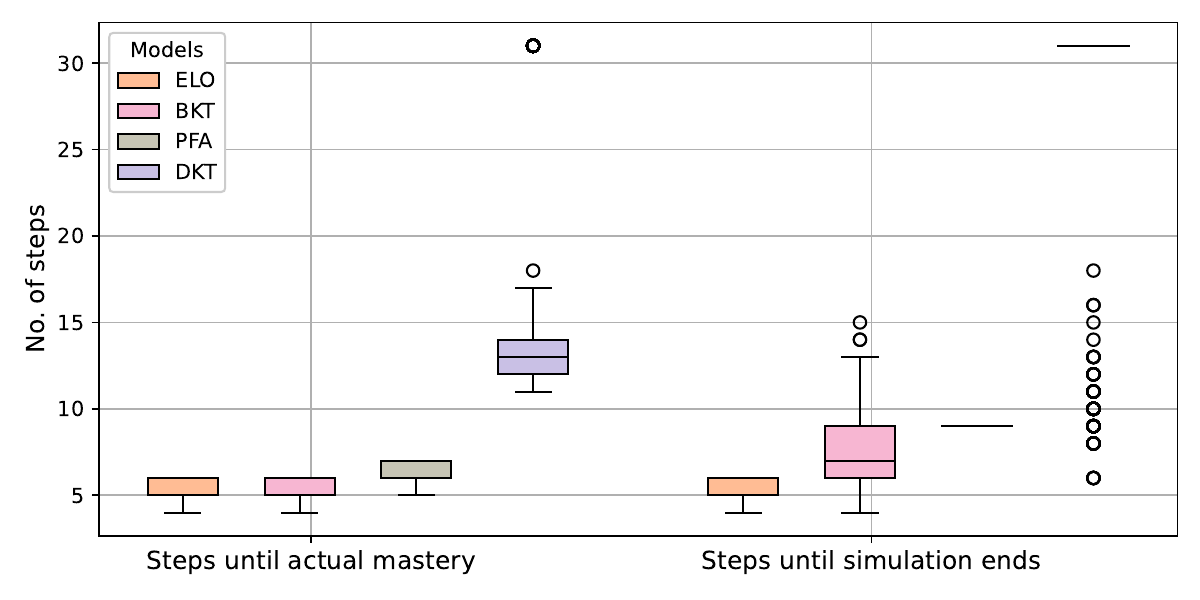}
\caption{The number of steps until mastery (left) and the number of steps until the simulation ends for the different KT models in the simulation study.}
\label{fig:steps_until_mastery}
\end{figure}

All models achieved reasonable model fit with test accuracies of 70.40\% for BKT, 81.04\% for PFA, and 71.89\% for DKT.

We observe that PFA and BKT only stopped teaching if mastery was achieved, whereas DKT stopped before achieving mastery in 24\% of cases. DKT was also unreliable in stopping when mastery was actually achieved: In the majority of simulations, we had to enforce an upper limit of 30 tasks because the simulation was caught in endless loops, meaning DKT never predicted a success probability high enough that it would indicate mastery. By contrast, PFA consistently stopped after 9 tasks and BKT mostly between 6-8 tasks -- close to the optimal achievable value of 5-6 (Fig.~\ref{fig:steps_until_mastery}, left).  Mastery was actually achieved in PFA and BKT after 6 steps, consistent with the optimum (the value for Elo), whereas DKT needed a median of 14 steps (Fig.~\ref{fig:steps_until_mastery}, right). This difference is strongly significant in a Wilcoxon signed rank test ($p < 10^{-10}$). In other words, our simulation clearly confirms our hypothesis: The non-interpretable KT model, DKT, leads to much less effective teaching decisions compared to the intepretable KT models, PFA and BKT.

\section{Teacher Study}

Next, we repeat our simulation study, but let human teachers make the pedagogical decisions, namely which task to select next and when to stop.

\subsection{Methods}
\label{sec:study_methods}

\begin{figure*}[t]

\centering
\includegraphics[width=0.9\textwidth, trim={47cm 0 0 0},clip]{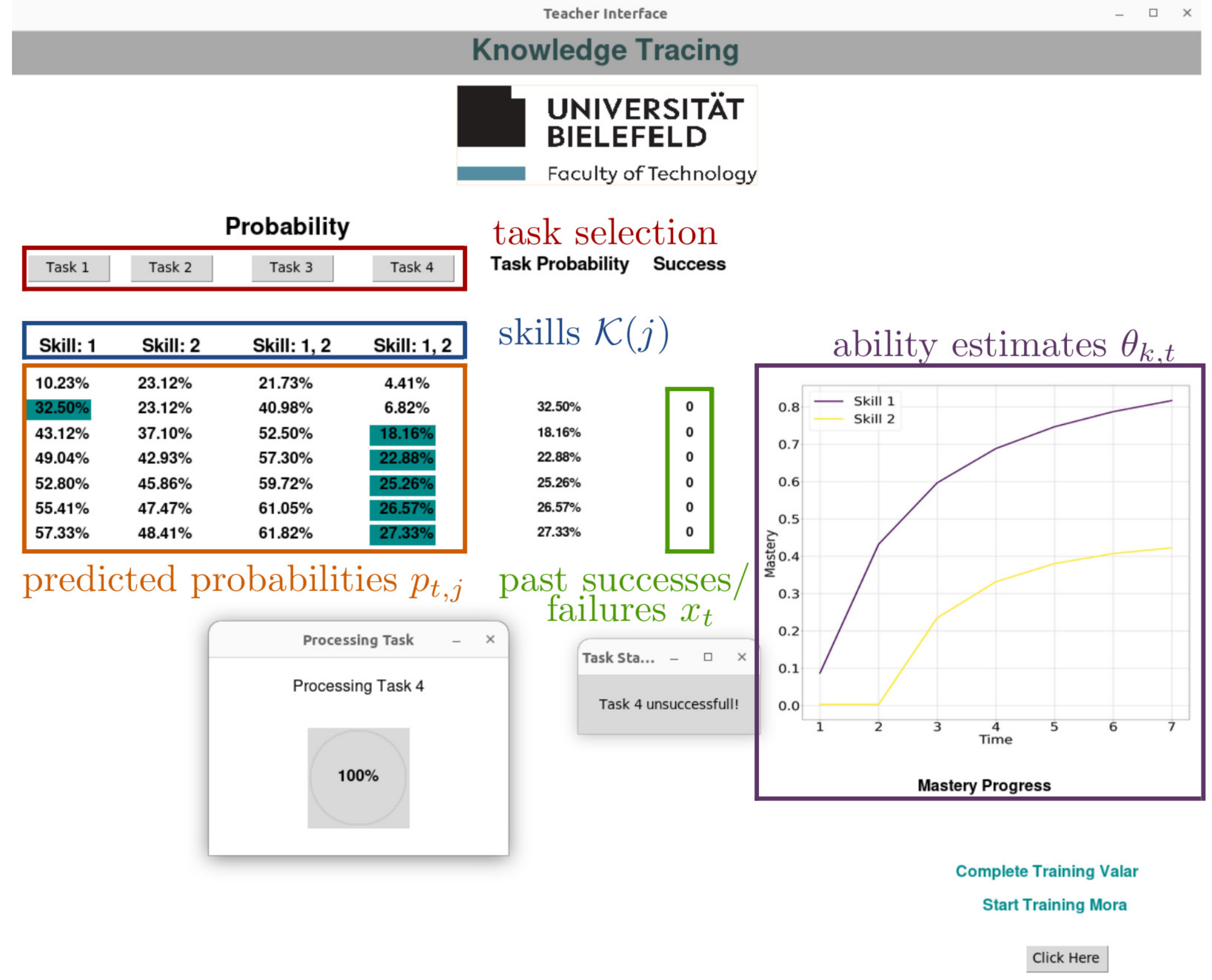}

\caption{The user interface for the teacher study. Teachers could select tasks (buttons on the top left) and decide when to stop (button on the bottom right) based on which skills belonged to which tasks, past successes and failures of the simulated student, predicted success probabilities on each task by the KT model, and the ability estimates of the KT model. Best viewed in color.}
\label{fig:interface}
\end{figure*}

\paragraph{User interface:} To enable teacher interaction with the simulation, we developed a user interface, shown in Fig.~\ref{fig:interface}. The interface was designed to facilitate two types of pedagogical decisions: which task the simulated student should attempt next (buttons on the top left), and when to stop teaching (button on the bottom right). Other than that, we did not want teachers to employ any domain knowledge about specific skills; therefore, we invoked a deliberately unrealistic domain, namely teaching camels space-traveling skills. Importantly, the user interface did present all information available based on the KT models, namely which task relates to which skill, the past successes and failures of the simulated student, the predicted success probabilities for each task by the current KT model, and (for BKT and PFA) a graph of the ability estimates.

\paragraph{Recruitment and ethics:} We recruited study participants among the teaching staff of an interdisciplinary research institute at a mid-sized central European university. Our recruitment criterion was a minimum teaching experience of 6 months. Participants had no prior knowledge about knowledge tracing models. Recruitment took place both via e-mail and in person. Overall, $N=12$ teachers were recruited. In line with European privacy regulations, to protect the privacy of participants, demographic data was not recorded and only anonymous data was kept for analysis. The study was approved by the local university ethics committee.
\paragraph{Protocol:} Participants were guided into a lab, received a briefing, and provided individual informed consent to participate in the study. Then, they were seated in front of a lab computer and performed nine teaching simulations in the user interface (Fig.~\ref{fig:interface}), three for BKT, three for PFA, and three for DKT. In each simulation, participants selected the next task for the simulated student by clicking one of four buttons on the top left of the interface, and could decide to stop teaching the current student by clicking the button on the bottom right. Participants did not know which underlying KT model was used in each simulation and the order of KT models was systematically varied across participants to cover all permutations. After each model, participants completed two questionnaires, namely the System Usability Scale (SUS) \cite{bangor2008sus} and the Trust of Automated Systems Test (TOAST) \cite{wojton2020toast} to evaluate the subjective usability and trustworthiness of each KT model. Our user interface additionally recorded all selected tasks $j_t$, the time until the next task selection, the sampled successes and failures $x_t$, all predicted probabilities $p_{t, j}$ and ability estimates $\theta_{t, k}$ of the respective KT model, as well as the ground truth probabilities $p^*_{t, j}$ and abilities $\theta^*_{t, k}$ of the Elo model.

\subsection{Results}
\label{sec:study_results}

\begin{figure}
\centering
\includegraphics[width=1\linewidth]{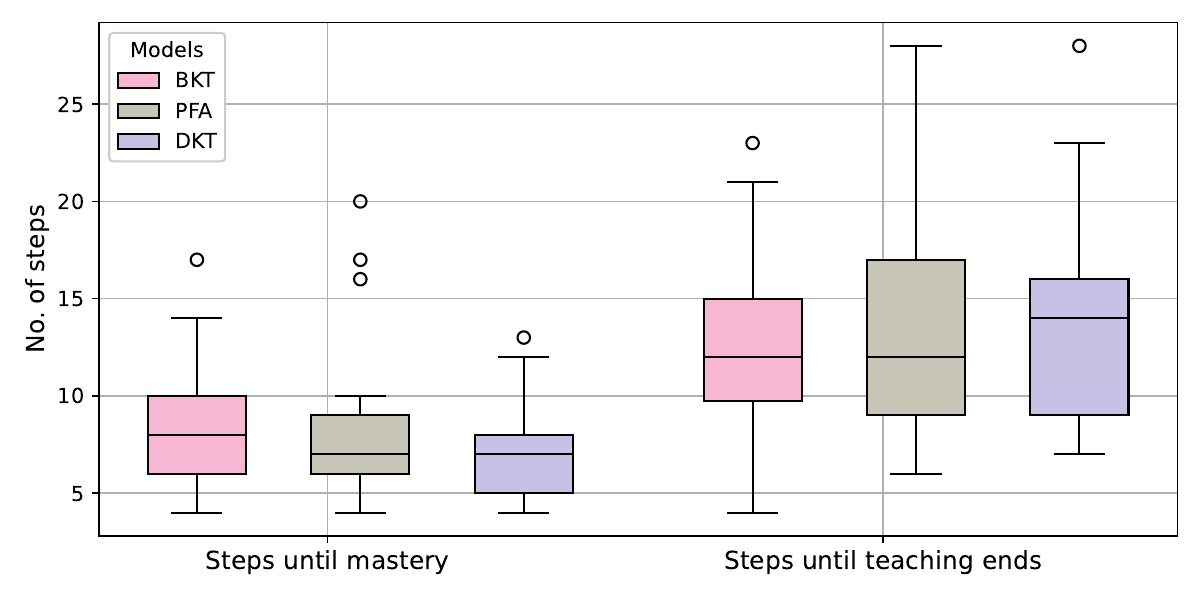}
\caption{The number of steps until mastery (left) and the number of steps until teaching ended (right) for the different KT models in the teacher study.}
\label{fig:steps_until_mastery_teachers}
\end{figure}

\begin{figure*}
\includegraphics[width=\linewidth]{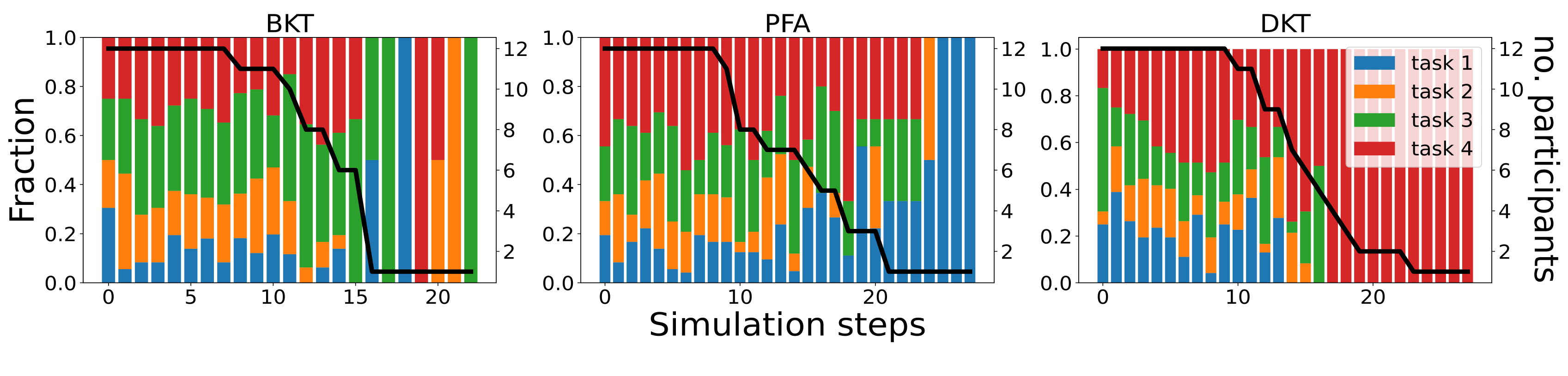}
\caption{Fraction of selected tasks across time for BKT (left), PFA (center), and DKT (right) across teachers in each step of the simulation.}
\label{fig:task_selection}
\end{figure*}

We first observe that teachers sometimes stopped teaching too early: In the BKT condition, 16\% of simulations were stopped before achieving mastery, in PFA 5\%, and in DKT 3\%. The number of simulation steps is shown in Fig.~\ref{fig:steps_until_mastery_teachers}. Teachers continued the simulation for 12 tasks in median for BKT and PFA, and for 14 tasks for DKT (the difference was not significant). However, counter to our hypothesis, the simulated students actually achieved mastery \emph{faster} in the DKT condition compared to BKT ($p \approx 0.047$ in a one-sided Welch $t$-test) and there was no significant difference between PFA and DKT (median of 8 tasks for BKT and 7 for PFA and DKT). This result can be explained with the tasks that were selected: In the DKT condition, teachers selected task 4 much more frequently, which was the optimal choice in our simulation (Fig.~\ref{fig:task_selection}). However, we emphasize that the user interface did not suggest this choice, indicating that teachers considered information beyond what was provided by KT models.

\paragraph{Decision Time:} Given that the interpretable models (BKT and PFA) provide more information for teachers to process, we would assume that they need more time to decide which task to select next. In fact, differences are very slight with all medians slightly below 7.5 seconds per decision, across conditions. Still, we detect significantly higher times in BKT compared to DKT ($p < 0.05$ in a $t$-test) and PFA compared to DKT ($p < 0.05$ in a $t$-test).

\paragraph{Usability and Trust Ratings:} Note that the user interface between conditions only differed in the graph displayed on the right side of the interface (Fig.~\ref{fig:interface}) with a graph for BKT and PFA and no graph for DKT. Teachers rated the usability of the BKT condition and the PFA condition (both ca. \ 60 points on the SUS scale) higher than DKT (ca.\ 50 points). The difference between BKT and DKT was significant in a $t$-Test ($p < 0.05$). In terms of the Trust of Automated Systems Test (TOAST), we observe significant differences in the understanding subscale of TOAST between BKT and DKT ($p < 0.01$, effect size $d=0.84$), and in both TOAST subscales between PFA and DKT ($p < 0.05$, effect sizes $d = 0.43$ and $d=0.81$), indicating that teachers trusted the interpretable KT models more.

\section{Limitations}
\label{sec:limitations}

All our results should be interpreted with several limitations in mind. First, our recruitment procedure only included university-level teachers with relatively little teaching experience on average. Results for a different teacher population (primary, secondary, vocational, or continued education) may be different.

Second, and perhaps most importantly, our simulation environment does not capture a realistic teaching setup. In our environment, teachers can only control which task to give to a student (corresponding to the outer loop of an intelligent tutoring system \cite{vanlehn2006behavior}) and they only receive the information whether a student was successful or not and what a KT model makes of that information. We deliberately used such a reduced laboratory setup to ensure a consistent learning behavior of the simulated students and to keep the amount of time needed for the study feasible. However, in the real world, teachers would receive much richer information, would have a much broader range of possible pedagogical decisions, and countless confounding variables would interfere with the effect of interpretability.

Third, even within the simulation environment, we had to make some design choices that may influence the results. It may be that for a different number of tasks, different difficulty parameters, different learning rates in the Elo model, a different ground-truth model, or a different user interface, different results would have occurred. The user interface is particularly noteworthy given that our participants gave rather low usability ratings, indicating the need for improvements. In more detail, it may well be that teachers had a hard time parsing the provided information correctly and that a user interface that better explained the provided information may have led to different outcomes. We encourage further research with variations of our setup and we share the source code for our study at \url{https://gitlab.ub.uni-bielefeld.de/publications-ag-kml/kt_space_tutor}.

\section{Discussion and Conclusion} \label{sec:conclusion}

In a simulation study and a teacher study, we investigated whether interpretable KT models (BKT and PFA) are more helpful in pedagogical decision making compared to a non-interpretable KT model (DKT). In particular, we investigated which task to select next for a simulated student and when to stop teaching. In our simulation study, BKT and PFA clearly outperformed DKT with less tasks needed until mastery and more reliable stopping decisions. However, when $N=12$ human teachers made the decisions, we did not observe any difference in number of tasks until stopping, and teachers selected slightly \emph{better} tasks in the DKT condition compared to the BKT and PFA condition. There may be multiple explanations for this finding: First, teachers may have been cognitively overloaded by the graphs and, hence, may have decided more accurately when the interface provided less information. This hypothesis is in slight tension with usability ratings, though: teachers preferred the interface the BKT and PFA conditions, indicating that they did not feel more overloaded in these conditions. Second, teachers may have overtrusted the graphs provided by BKT and PFA and, therefore, may have been mislead by these graphs. This explanation is in line with the observation that teachers tended to stop slightly too early in the BKT condition which also tended to provide overly optimistic mastery estimates in the graph. However, it does not explain the precise task choices teachers made and it does not explain why teachers did not overtrust the DKT model -- which would predict even poorer task choices according to our simulation. Therefore, a third explanation is that teachers may have mis-interpreted the graphs provided by BKT and PFA, leading them to select sub-optimal tasks (especially tasks 1 and 2, which only targeted a single skill).

We believe that this suggests ample opportunity for future work, namely a) improved KT models that retain the desirable interpretability properties of PFA and BKT while being more accurate, b) better user interfaces for and explanations of KT models to help teachers in pedagogical decision making, and c) more research into how learners and teachers actually use KT model information and how pedagogical decision making can actually be improved based on KT models.

\section*{Acknowledgement}

We thank the reviewers for their comments that informed our discussion of findings.
Part of this work has been part of the project \enquote{Explaining Learner Models using Language Models} as part of KI:edu.nrw. KI:edu.nrw is funded by the Ministry of Culture and Science of the State of North Rhine-Westphalia. We gratefully acknowledge this support.

\bibliographystyle{plainnat}
\bibliography{literature}

\end{document}